\documentclass{llncs}
\usepackage{multirow}
\usepackage{caption}
\usepackage{booktabs}
\usepackage{epsfig}
\usepackage{amsfonts}
\usepackage{longtable}
\usepackage{rotating}
\graphicspath{{fig/}}
\usepackage{color}

\usepackage{amsmath,amsfonts,amssymb}

\newcommand{\ve}[1]{\mathbf{#1}}

\begin{document}
\frontmatter          

\title{Cross-view Relation Networks \\ for Mammogram Mass Detection}
%

\author{Jiechao Ma\inst{1,3} \and Sen Liang\inst{3} \and Xiang Li\inst{1} \and Hongwei Li\inst{2} \and Bjoern H Menze\inst{2}  \and Rongguo Zhang \inst{3}\and Wei-Shi Zheng\inst{1}}
\authorrunning{J. Ma et al.} 
\institute{Sun Yat-sen University, Guangzhou, China,\\
\email{majch7@mail2.sysu.edu.cn}
\and
Technical University Munich, Munich, Germany\\
\and 
Infervision Inc. , Beijing, China,\\
}

\maketitle              

\begin{abstract}

{Mammogram is the most effective imaging modality for the mass lesion detection of breast cancer at the early stage. The information from the two paired views (i.e., medio-lateral oblique and cranio-caudal) are highly relational and complementary, and this is crucial for doctors' decisions in clinical practice. However, existing mass detection methods do not consider jointly learning effective features from the two relational views. To address this issue, this paper proposes a novel mammogram mass detection framework, termed Cross-View Relation Region-based Convolutional Neural Networks (CVR-RCNN). The proposed CVR-RCNN is expected to capture the latent relation information between the corresponding mass region of interests (ROIs) from the two paired views. Evaluations on a new large-scale private dataset and a public mammogram dataset show that the proposed CVR-RCNN outperforms existing state-of-the-art mass detection methods. Meanwhile, our experimental results suggest that incorporating the relation information across two views helps to train a superior detection model, which is a promising avenue for mammogram mass detection. }

\end{abstract}
\section{Introduction}
Breast cancer is by far the most common cancer diagnosed in women worldwide. In clinical practice, contextual information and multi-view information (i.e., medio-lateral oblique (MLO) view which is a side view of the breast taken at a certain angle, and cranio-caudal (CC) view which is a top-bottom view of the breast) are helpful for the radiologists to detect mass on the mammogram.
However, while there has been a significant progress in mass detection~\cite{Pengcheng} and classification~\cite{Wentao} based on mammogram by using deep convolutional neural networks (CNNs), most of the deep convolutional architectures identify the mammogram mass without taking different views into account during model training, such that the relation between two views of the mammogram cannot be learned.  

To address this limitation, in this work, we mainly focus on the aggregation of two views. Breast lesions could be at arbitrary image locations, of different scales, and from different categories. This makes it difficult to \emph{directly} model the related mass between two views of mammogram images. To solve this issue, we turn attention to the regions of interest (ROIs, i.e., candidate mass regions) detected by the region-based convolutional neural network architecture (RCNN), and explore the hidden relationship between the ROIs from two views. 
In particular, we propose a novel mammogram mass detection framework, termed cross-view relation region-based convolutional neural network (CVR-RCNN). Unlike the previous deep learning work \cite{Pengcheng,Wentao} which do not distinguish different views of mammogram, we extend a two-branch Faster RCNNs by including a novel cross-view relation network. In particular, we demonstrate the benefit of incorporating the relation information between the different views in our framework. 


Our contributions are twofold. First, to the best of our knowledge, this is the \textit{first} work to exploit the modeling relation information between two views for mammogram mass detection. Our cross-view detection framework is much more effective and efficient than existing approaches. Second, we introduce a new cross-view relation network for mass ROIs interaction modeling which imitates the flow of radiologists screening mammogram targets. Since the current datasets are not large enough, we collected a large-scale dataset that contains 1,425 specimen mammograms with annotations of breast masses and evaluated the proposed model on this challenging dataset. Public Digital Database for Screening Mammography (DDSM) dataset was also used to additionally justify the effectiveness of the proposed model. Our experimental results show that the proposed model outperforms several state-of-the-art methods. 

In principle, our approach is fundamentally differs from the previous mammogram mass detection methods. It is the \emph{first} end-to-end cross-view modeling framework using two-branch RCNNs, which opens a new avenue for mammogram mass detection and may further reduce radiologists’ screening time.

\section{The CVR-RCNN Framework}
%


The proposed CVR-RCNN detection framework consists of a two-branch extended Faster RCNNs and a novel relation network.
 
\begin{figure}
	\begin{center}
		\includegraphics[width=\linewidth,height=0.33\linewidth]{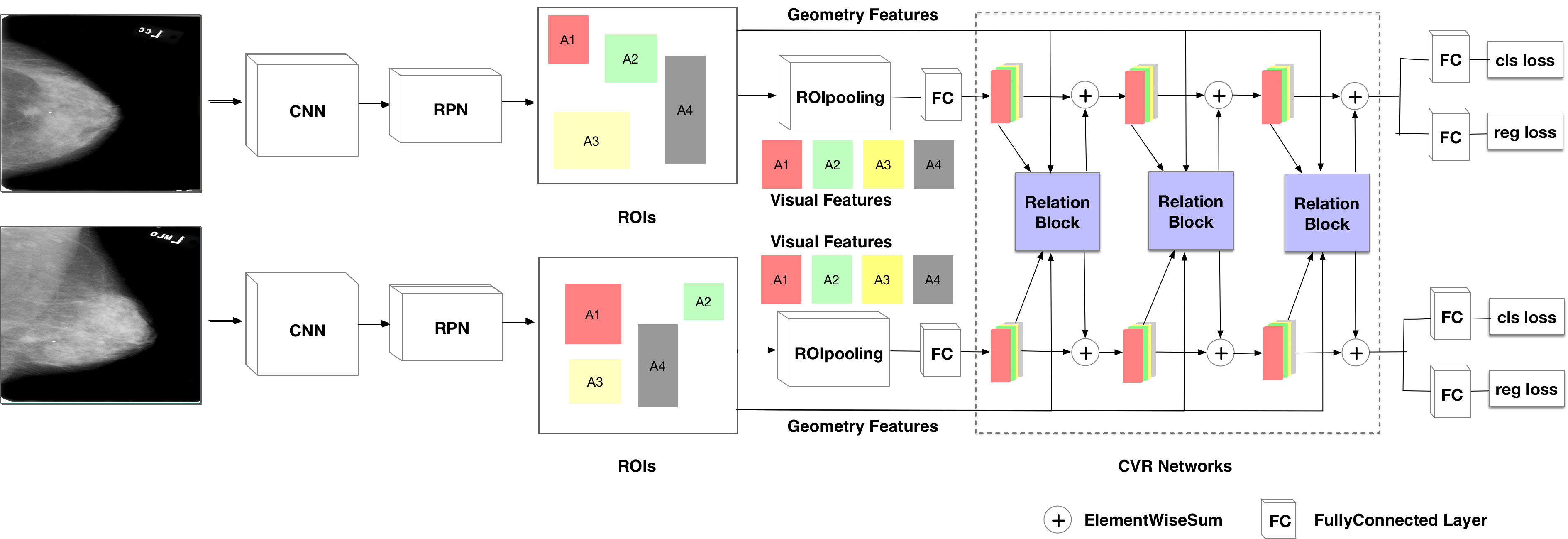}
	\end{center}
	\caption{The architecture of our CVR-RCNN framework. A paired input image set is fed into the two-branch Faster RCNNs to get the ROIs. The visual and geometry features of ROIs are used by cross-view relation network to learn the effective relation features. The designed losses are used to optimize the whole framework.}
	\label{fig:cvr} 
\end{figure}

\noindent\textbf{Two-branch Faster RCNNs.} 
Motivated by the siamese network \cite{zagoruyko2015learning} which uses two weight-shared feature extraction branches, in our framework, we propose a two-branch weight-shared RCNNs (Figure~\ref{fig:cvr}, Left half) connected by relation blocks to learn the latent cross-view information. For each branch, we adopt the current popular object detection framework Faster-RCNN \cite{ren2015faster} and extend it with several residual blocks. Further, inspired by the relevant work~\cite{he2016identity}, each original residual block~\cite{he2016deep} was modified by performing batch normalization (BN) and rectified linear unit (ReLU) before convolution operations. Each branch of Faster RCNN aims to detect the regions of interest (ROIs, i.e., the candidate mass regions) for further processing by the following cross-view relation network.


%
%

\noindent\textbf{Cross-view Relation Networks.} 
In order to discover the latent relation between two views, inspired by \textbf{relationship module} proposed in \cite{hu2018relation}, we designed a new relation network consisting of certain number of \textbf{relation blocks} linking the paired views of mammogram (see Figure~\ref{fig:cvr}, `CVR Networks'). The objective of the relation network is to transfer both visual and geometric information of ROIs from the second (or first) view to the first (or second) one in order to help detect masses more effectively in the first (or second) view.  

Suppose the $n^{th}$ ROI in the first view needs to be classified and its position needs to be fine-tuned from the first branch of the proposed framework. Denote the visual feature representation of the  $n^{th}$ ROI in the first view by $\ve{f}_{1,n}$, and the geometric feature by $\ve{p}_{1,n}$. In general, $\ve{f}_{1,n}$ is a feature vector from the output of a fully connected layer following the the ROI-pooling layer in the Faster RCNN for the $n^{th}$ ROI region, and $\ve{p}_{1,n} = [x_{1,n}, y_{1,n}, w_{1,n}, h_{1,n}]^\intercal$ includes the coordinate ($x_{1,n}, y_{1,n}$), width ($w_{1,n}$), and height ($h_{1,n}$) of the  $n^{th}$ ROI in the first view. Similarly, denote the  visual feature representation of the  $m^{th}$ ROI in the second view by $\ve{f}_{2,m}$, and its geometric feature by $\ve{p}_{2,m} = [x_{2,m}, y_{2,m}, w_{2,m}, h_{2,m}]^\intercal$.

In order to use the visual information from the second view to help detect the $n^{th}$ mass candidate (represented by $\ve{f}_{1,n}$ and $\ve{p}_{1,n}$) in the first view, we need to establish both the visual and geometric relations between each mass candidate in the second view and the  $n^{th}$ candidate in the first view. The strength of visual relation or the similarity between the $n^{th}$ candidate in the first view and the $m^{th}$ candidate in the second view can be represented by
\begin{equation}
w_{n,m} = \frac{(\ve{W}_1 \ve{f}_{1,n}) ^\intercal(\ve{W}_2 \ve{f}_{2,m})}{\sqrt{d}} \,,
\end{equation}
where $\ve{W}_1$ and $\ve{W}_2$ are two matrices transforming the original visual features into the same feature space before measuring the similarity based their dot product. $\ve{W}_1$ and $\ve{W}_2$ are part of the model parameters to be learned. $d$ is the dimension of the new feature space and $\sqrt{d}$ is used as a normalization factor.

To establish geometric relationship between the candidates in the two views, inspired by the work~\cite{girshick2014rich}, we first normalize the geometric information of the $n^{th}$ candidate in the first view by the $m^{th}$ candidate in the second view, 
\begin{equation}
\ve{g}_{n,m} =[\log(\frac{\left |x_{1,n} -x_{2,m} \right |}{w_{2,m}}), \log(\frac{\left |y_{1,n} -y_{2,m} \right |}{h_{2,m}}), \log(\frac{w_{1,n}}{w_{2,m}}), \log(\frac{h_{1,n}}{h_{2,m}})]^\intercal \,,
\end{equation}
where $\log(\cdot)$ is used to boost potential geometric relation between candidates by reducing the effect of differences in position and size between the two candidates. Then similar to the work~\cite{hu2018relation,vaswani2017attention}, the cross-view normalized geometric feature $\ve{g}_{n,m}$ is embedded to a high-dimensional feature space by the $\sin(\cdot)$ and $\cos(\cdot)$ functions of the $\ve{g}_{n,m}$ elements at different frequencies (please refer to the reference~\cite{vaswani2017attention} for details). The embedding process is denoted by $\ve{E}(\ve{g}_{n,m})$. The geometric relation between two candidates is then defined as 
\begin{equation}
v_{n, m} = \max(0, \ve{v}^\intercal \ve{E}(\ve{g}_{n,m})) \,,
\end{equation}
where $\ve{v}$ is a vector transforming the high-dimensional feature vector $\ve{E}(\ve{g}_{n,m})$ into a scalar weight. $\ve{v}$ is part of the model parameters to be learned. The $\max({0, \cdot})$ function is used to trim any negative weight $\ve{v}^\intercal \ve{E}(\ve{g}_{n,m})$ to $0$, thus restricting interaction between candidates satisfying certain geometric relationships.

Combining the visual and geometric information, the relation between the $n^{th}$ candidate in the first view and all the candidates in the second view can be summarized as
\begin{equation}
\ve{f}'_{1,n} = \frac{1}{\sum_{k} v_{n, k} \exp{(w_{n,k})}}\sum_{m} v_{n, m} \exp{(w_{n,m})} \cdot \ve{W}_3 \ve{f}_{2,m} \,.
\end{equation}
Here $\exp(\cdot)$ is used to make sure the visual relation is strengthened and non-negative between similar candidates (as in the reference~\cite{hu2018relation}), and $\sum_{k} v_{n, k} \exp{(w_{n,k})}$ is a normalization factor. Each visual feature $\ve{f}_{2,m}$ from the second view is transformed by the matrix $\ve{W}_3$ to the feature space in which the $\ve{f}_{1,n}$ is. Since $\ve{f}_{1,n}$ and $\ve{f}_{2,m}$ are often extracted similarly from corresponding ROIs,  $\ve{W}_3$ is in general a square matrix. $\ve{W}_3$ is also part of the model parameters to be learned.

Finally, the relational feature $\ve{f}'_{1,n}$ is added to the original feature $\ve{f}_{1,n}$ to form the output of the relation block for the $n^{th}$ mass candidate in the first view,
\begin{equation}
\tilde{\ve{f}}_{1,n} = \ve{f}_{1,n} + \ve{f}'_{1,n} \,.
\end{equation}
Motivated by the skip connection in the ResNet~\cite{he2016deep}, the two visual features $\ve{f}_{1,n}$ and $\ve{f}'_{1,n}$ are summed rather than concatenated. The output of the relation block $\tilde{\ve{f}}_{1,n}$ (and the original geometric feature $\ve{p}_{1,n}$) can be used as input to the next relation block when multiple relation blocks are employed in the detection framework.

For each ROI from the first (or second) view, the output from the last cross-view relation block was then fed into a number of fully connected layers for ROI classification and another number of fully connected layers for bounding box offset prediction, as in the Faster RCNN. Given a set of paired (two-view) images, the two-branch Faster RCNNs with the cross-view relation network can be trained by minimizing the loss function $L$,
\begin{equation}
L = L_{1,cls} + \alpha  L_{1,reg} +  \beta L_{2,cls} + \gamma L_{2,reg}  \,,
\end{equation}
where $L_{i,cls}$ and $L_{i,reg}$ respectively represent the classification and the regression losses from the $i^{th}$ view. $\alpha, \beta, \gamma$ are coefficients balancing the loss terms. 

%
%

\section{Experiments}
\subsection{Experimental Protocols}
\noindent\textbf{Private Dataset}: To evaluate the proposed framework on the large breast mass dataset, we collected a large-scale dataset which contains 1,425 scanned mammography images with breast mass lesions. To the best of our knowledge, this dataset is the largest cohort collected specifically for breast masses detection. The annotations (bounding boxes of each breast lesion) were labeled by 4 experienced radiologists. Each senior radiologist evaluated the annotations made by the relatively junior experienced radiologist and made further modifications, if necessary.

\noindent\textbf{DDSM Dataset}: DDSM is a publicly available mammography dataset. As in other studies \cite{campanini2004novel,eltonsy2007concentric,sampat2008model}, 512 mammography films were used in the evaluation.

\noindent\textbf{Data Preprocessing}: To avoid over-fitting during training, the training set was augmented by affine transformations (e.g., rotations). Each image was resized to $1024\times1024$ pixels and
the pixel values were rescaled to the range $[0, 255]$.

\noindent\textbf{Training Details}:
MXNET library was applied to construct the proposed deep convolutional architecture. 
The model parameters were initialized by a pretrained ResNet-101 and then fine-tuned around 20 epochs using the early-stopping criterion with a mini-batch of two images for each device with 4 GPUs. The SGD optimizer was used with learning rate $0.001$. And we evaluated the proposed framework with default setting ($N=3$, $\alpha = 2, \beta = 1, \gamma = 2$).

\noindent\textbf{Statistical Evaluation}: In both datasets, about $80\%$ paired images were used for training, and the remaining $20\%$  for testing. $F_1$ score, precision and recall were used as evaluation metrics. For the public dataset, true positive rate (TPR) versus false positive per image (FPI) were used as the metrics following previous work (e.g., \cite{sampat2008model}).

\subsection{Effect of the Relation Block}
In this section, the effect of the number of relation blocks in the cross-view relation network was investigated based on our private dataset.   
As shown in Table 1, adding the relation blocks to the network (second to last rows) clearly improved the detection performance than that without relation block (first row, $N=0$). Also, more relation blocks steadily lead to the higher precision rate (e.g., achieving 76.56$\%$ when using $4$ relation blocks). However, compared with the higher precision rate, the recall is largely reduced when $4$ relation blocks were used (clinically, recall is relatively more important than precision). One reason could be that by sharing features and relation more times between the two views, the network is forced to pay more attention to the relationship between the two views, rather than to the visual features from each single view. In addition, using more relation blocks would increase the computational complexity and memory usage. As a trade-off, $3$ relation blocks were used as default in the tests below. 

\begin{table}{}
	\newcommand{\tabincell}[2]{\begin{tabular}{@{}#1@{}}#2\end{tabular}}
	\renewcommand\arraystretch{1}
	\centering
	\caption{Effect of relation block(s) in the cross-view relation network on our private dataset. $N = 0$ corresponds to the two-branch Faster RCNNs without relation blocks.}
	\begin{tabular}{lcccc}
		\toprule
		
		\textbf{Relation Block($N$)}&\textbf{Precision(\%)}&\textbf{Recall(\%)}&\textbf{$F_1$ Score}&\textbf{FPI}\\
		\midrule
		$N$=0&65.27 & 71.93 & 0.69 & 0.42 \\
		$N$=1&69.66 & 71.70 & 0.71 & 0.35 \\
		
		$N$=2&70.10&72.13&0.71&0.33 \\
		
		$N$=3&71.12&\textbf{75.33}&\textbf{0.73} &0.30 \\
		$N$=4 & \textbf{76.56} & 70.39 & \textbf{0.73} & \textbf{0.27} \\		
		\bottomrule
	\end{tabular}
	\label{table:Table4}
\end{table}

\subsection{Comparison with Non Cross-view Methods} 
The proposed CVR-RCNN model was further evaluated on our private dataset by comparing with two representative detection frameworks Faster RCNN~\cite{ren2015faster} and SSD~\cite{Liu2016SSD}.
In Table~\ref{table:baselines}, `Faster RCNN' and `SSD' indicate that the data from two views are mixed (therefore not using view information) for training and testing. In comparison, `two-branch Faster RCNNs' and `two-branch SSDs' indicate that each view of data was used to train a individual detection model, and each test data was predicted by either the first or the second model based on which view it is from. As shown in Table~\ref{table:baselines}, the two-branch Faster RCNNs and SSDs models perform clearly better than their corresponding versions without considering view information, e.g., for the two-branch Faster RCNNs model, the precision rate 65.27\% vs. 64.01\% and the recall rate 71.93\% vs. 70.53\%.
This suggests that the separated models conditioned on views information outperforms the single model trained on the mix-view images.

\begin{table}{} 	
	\newcommand{\tabincell}[2]{\begin{tabular}{@{}#1@{}}#2\end{tabular}}
	\renewcommand\arraystretch{1}
	\centering
	\caption{Comparisons with different detection models on our private dataset.} 
	\begin{tabular}{lcccc}
		\toprule
		
		\textbf{Methods}&\textbf{Precision(\%)}&\textbf{Recall(\%)}&\textbf{$F_1$ Score}&\textbf{FPI}\\
		\midrule
		
		Faster RCNN~\cite{ren2015faster} &64.01&70.53&0.67&0.45 \\
		
		two-branch Faster RCNNs &65.27&71.93&0.69&0.42 \\
		
		SSD~\cite{Liu2016SSD} & 65.75 & 66.91 & 0.66 & 0.42\\
		
		two-branch SSDs & 66.40 & 68.30 & 0.67 & 0.41\\
		
		\midrule
		
		\textbf{CVR-RCNN} &\textbf{71.12}&\textbf{75.33} &\textbf{0.73}&\textbf{0.30}\\
		
		\bottomrule
	\end{tabular}
	\label{table:baselines}
\end{table}

However, for the two-branch Faster RCNNs model, it did not consider that the lesions in both of the two views may be largely related to each other. By learning the relationship between the two views, using the \textit{Cross-View Relation Network}, our CVR-RCNN achieved the best performance, producing the notable improvements of the precision rate from 65.27$\%$ to 71.12$\%$, and the recall rate from 71.93\% to 75.33\%. One reason for the notable improvement should be that, by sharing the visual and geometric information of the ROIs between the two views, the network is driven to learn the different manifestations of the same lesions in two views. Then for each ROI, the network would examine whether the ROI exists in both views. After such double check strategy, the false positive rate would be likely reduced, improving the precision rate and reducing the false positives per image (FPI).


\begin{table}{}
	\newcommand{\tabincell}[2]{\begin{tabular}{@{}#1@{}}#2\end{tabular}}
	\renewcommand\arraystretch{1}
	\centering
	\caption{Comparisons with state-of-the-art methods with true positive rate (TPR) versus FPI on the public DDSM dataset.}
	\begin{tabular}{lcccc}
		\toprule
		
		\textbf{Methods}&\textbf{$F_1$ Score}&\textbf{TPR@FPI}\\
		\midrule
		Campanini et al.~\cite{campanini2004novel}&-&0.80@1.1 \\
		
		Eltonsy et al.~\cite{eltonsy2007concentric} &-&0.92@5.4, 0.88@2.4, 0.81@0.6 \\
		
		Sampat et al.~\cite{sampat2008model}&-&0.88@2.7, 0.85@1.5, 0.80@1.0 \\
		\midrule
		
		Faster RCNN~\cite{ren2015faster} &0.52 &0.85@2.1, 0.75@,1.8, 0.73@1.2\\
		two-branch Faster RCNNs & 0.57&0.75@1.0, 0.73@0.9\\

		\textbf{CVR-RCNN} &\textbf{0.75}&\textbf{0.92@4.4, 0.88@1.9, 0.85@1.2} \\
		\bottomrule
	\end{tabular}
	\label{table:leaderboard}
\end{table}

\subsection{Comparison with State-of-the-Art} 
To further justify the effectiveness of the proposed model, we performed experiments on the public DDSM dataset and compared with the results from the public DDSM leaderboard. Table~\ref{table:leaderboard} shows that our proposed method clearly outperforms the state-of-the-art methods for mass detection in mammograms, where the results from the competing methods are reported by their original authors. The results demonstrate that the proposed CVR-RCNN is noticeably better than the previous ones, suggesting that our CVR-RCNN is more suitable for mammogram mass detection.

In addition, the results from the Faster RCNN and the two-branch Faster RCNNs (Table~\ref{table:leaderboard}, fourth and fifth rows) again confirm that view information is helpful to improve the detection performance, with better performance from the two-branch Faster RCNNs. More importantly, by comparing the results of the two-branch Faster RCNNs and the proposed CVR-RCNN, we can see that the model trained with cross-view relation network obtain the best performance, demonstrating that the interaction between two views is effective for detecting breast masses in the proposed framework.


\section{Conclusion}

In this work, we proposed a cross-view mammogram mass detection framework by combining the conventional CNN detection with a novel cross-view relation network. Extensive evaluations on a private dataset and a public dataset clearly demonstrated the superior performance of the proposed framework compared to state-of-the-art methods. This opens a new avenue to improve detection of mammogram mass where paired or multiple views of information are available.


{
	\bibliographystyle{splncs03}
	\bibliography{egbib}
}

\end{document}